\documentclass{article}

\input{Config/packages}
\glsdisablehyper

\newacronym{adse}{ADSE}{Automatic Design Space Exploration}
\newacronym{aer}{AER}{Address Event Representation}
\newacronym{ai}{AI}{Artificial Intelligence}
\newacronym{ann}{ANN}{Artificial Neural Network}
\newacronym{api}{API}{Application Programming Interface}
\newacronym{asic}{ASIC}{Application-Specific Integrated Circuit}
\newacronym{ax}{AX}{Adaptive eXperimentation}
\newacronym{bptt}{BPTT}{Back-Propagation Through Time}
\newacronym{bram}{BRAM}{Block RAM}
\newacronym{cl}{CL}{Continual Learning}
\newacronym{cnn}{CNN}{Convolutional Neural Network}
\newacronym{cots}{COTS}{Commercial Off The Shelf}
\newacronym{cpu}{CPU}{Central Processing Unit}
\newacronym{csnn}{CSNN}{Convolutional Spiking Neural Network}
\newacronym{cu}{CU}{Control Unit}
\newacronym{dnn}{DNN}{Deep Neural Network}
\newacronym{dp}{DP}{Data Path}
\newacronym{dram}{DRAM}{Dynamic Random Access Memory}
\newacronym{dse}{DSE}{Design Space Exploration}
\newacronym{dvs}{DVS}{Dynamic Vision Sensor}
\newacronym{eda}{EDA}{Electronic Design Automation}
\newacronym{elut}{ELUT}{Equivalent Look Up Table}
\newacronym{eth}{ETH}{Eidgenössische Technische Hochschule}
\newacronym{fc}{FC}{Fully-Connected}
\newacronym{fcr}{FC-R}{Fully-Connected Recurrent}
\newacronym{ff}{FF}{Flip Flop}
\newacronym{fffc}{FF-FC}{Feed-Forward Fully-Connected}
\newacronym{FFNN}{FFNN}{Feed-Forward Neural Network}
\newacronym{fi}{FI}{Fault Injection}
\newacronym{fim}{FIM}{Fault Injection Manager}
\newacronym{fl}{FL}{Fault List}
\newacronym{flg}{FLG}{Fault List Generator}
\newacronym{fpga}{FPGA}{Field Programmable Gate Array}
\newacronym{fsm}{FSM}{Finite State Machine}
\newacronym{gpgpu}{GPGPU}{General Purpose Graphic Processing Unit}
\newacronym{gpu}{GPU}{Graphic Processing Unit}
\newacronym{hdl}{HDL}{Hardware Description Language}
\newacronym{hw}{HW}{Hardware}
\newacronym{if}{IF}{Integrate and Fire}
\newacronym{ini}{INI}{Institute of Neuro-Informatics}
\newacronym{iot}{IoT}{Internet of Things}
\newacronym{lif}{LIF}{Leaky Integrate and Fire}
\newacronym{lr}{LR}{Latent Replay}
\newacronym{lstm}{LSTM}{Long Short Term Memory}
\newacronym{lut}{LUT}{Look Up Table}
\newacronym{mac}{MAC}{Multiply and Accumulate}
\newacronym{ml}{ML}{Machine Learning}
\newacronym{mlp}{MLP}{Multi-Layer Perceptron}
\newacronym{nas}{NAS}{Network Architecture Search}
\newacronym{ne}{NE}{Network Evaluator}
\newacronym{ng}{NG}{Network Generator}
\newacronym{nir}{NIR}{Neuromorphic Intermediate Representation}
\newacronym{nlp}{NLP}{Natural Language Processing}
\newacronym{nn}{NN}{Neural Network}
\newacronym{ostl}{OSTL}{Online Spatio-Temporal Learning}
\newacronym{ostp}{OSTP}{Online Spatio-Temporal Learning with Target Projection}
\newacronym{pu}{PU}{Processing Unit}
\newacronym{pulp}{PULP}{Parallel processing Ultra-Low Power}
\newacronym{qat}{QAT}{Quantization Aware Training}
\newacronym{ram}{RAM}{Random Access Memory}
\newacronym{rcr}{RC-R}{Randomly-Connected Recurrent}
\newacronym{rl}{RL}{Reinforcement Learning}
\newacronym{rnn}{RNN}{Recurrent Neural Network}
\newacronym{rom}{ROM}{Read Only Memory}
\newacronym{rsnn}{RSNN}{Recurrent Spiking Neural Network}
\newacronym{rtl}{RTL}{Register Transfer Level}
\newacronym{rtrl}{RTRL}{Real-Time Recurrent Learning}
\newacronym{sbs}{SbS}{Spike-by-Spike}
\newacronym{scnn}{SCNN}{Spiking Convolutional Neural Networks}
\newacronym{sfi}{SFI}{Statistical Fault Injection}
\newacronym{sg}{SG}{Surrogate Gradient}
\newacronym{shd}{SHD}{Spiking Heidelberg Dataset}
\newacronym{simd}{SIMD}{Single Instruction Multiple Data}
\newacronym{snn}{SNN}{Spiking Neural Network}
\newacronym{soc}{SoC}{System on Chip}
\newacronym{sota}{SOTA}{State Of The Art}
\newacronym{sram}{SRAM}{Static Random Access Memory}
\newacronym{srm}{SRM}{Spike Response Model}
\newacronym{stdp}{STDP}{Spike-Timing-Dependent Plasticity}
\newacronym{tpu}{TPU}{Tensor Processing Unit}
\newacronym{ucb}{UCB}{Upper Confidence Bound}
\newacronym{vhdl}{VHDL}{VHSIC Hardware Description Language}
\newacronym{wta}{WTA}{Winner Takes All}
\newacronym{ANN}{ANN}{Artificial Neural Network}
\newacronym{PNN}{PNN}{Photonic Neural Network}
\newacronym{ADC}{ADC}{Analog-to-Digital Converter}
\newacronym{DAC}{DAC}{Digital-to-Analog Converter}
\newacronym{MZI}{MZI}{Mach-Zehnder Interferometer}
\newacronym{MVM}{MVM}{Matrix-Vector Multiplication}
\newacronym{CMOS}{CMOS}{Complementary Metal-Oxide-Semiconductor}
\newacronym{ASIC}{ASIC}{Application-Specific Integrated Circuit}
\newacronym{SVD}{SVD}{Singular Value Decomposition}
\newacronym{epsp}{EPSP}{Excitatory Post-Synaptic Potential}
\newacronym{ltp}{LTP}{Long-Term Potentiation}
\newacronym{ltd}{LTD}{Long-Term Depression}
\newacronym{mzi}{MZI}{Mach-Zehnder Interferometers}
\newacronym{ptp}{PTP}{Post-Tetanic Potentiation}
\newacronym{bp}{BP}{Back-Propagation}
\newacronym{shl}{SHL}{Supervised Hebbian Learning}
\newacronym{hh}{HH}{Hodgkin–Huxley}
\newacronym{ip}{IP}{Intellectual Property}

\title{SFATTI: Spiking FPGA Accelerator for Temporal Task-driven Inference - A Case Study on MNIST\thanks{This work was partially supported by project SERICS (PE00000014) under the MUR National Recovery and Resilience Plan funded by the European Union.}}

\name{Alessio Caviglia, Filippo Marostica, Alessio Carpegna, Alessandro Savino, Stefano Di Carlo}
  
 \address{$^{\star}$ Politecnico di Torino, \\ 
 Control and Computer Engineering Department, Torino, Italy, \\
  e-mail:\{firstname.lastname\}@polito.it }

\begin{document}
%
\maketitle
\begin{abstract}
Hardware accelerators are essential for achieving low-latency, energy-efficient inference in edge applications like image recognition. \glspl{snn} are particularly promising due to their event-driven and temporally sparse nature, making them well-suited for low-power \gls{fpga}-based deployment. This paper explores using the open-source \texttt{Spiker+} framework to generate optimized \glspl{snn} accelerators for handwritten digit recognition on the MNIST dataset. \texttt{Spiker+} enables high-level specification of network topologies, neuron models, and quantization, automatically generating deployable HDL. We evaluate multiple configurations and analyze trade-offs relevant to edge computing constraints.

\glsresetall
\end{abstract}

\section{Introduction}
\label{sec:intro}
Edge computing devices are increasingly leveraging \glspl{ann} to perform image processing directly at the data source. By enabling local inference, these systems reduce latency, lower energy consumption, and minimize dependence on remote servers. Processing visual data on-device not only enhances privacy but also supports real-time decision-making, key requirements for applications constrained by tight power budgets and timing deadlines. 

Among various \gls{ann} paradigms, \glspl{snn} have emerged as particularly promising for edge applications due to their biologically inspired, event-driven computation \cite{li2024brain}. Unlike traditional \glspl{ann} that rely on continuous activations and dense matrix multiplications, \glspl{snn} encode and process information as discrete spike events occurring over time, mimicking the sparse and temporally asynchronous communication of biological neurons.
\glspl{snn} are well-suited for dynamic, temporally rich data due to their energy-efficient, sparse, and event-driven computation. They reduce memory and compute demands, making them ideal for audio processing, image processing, gesture recognition, and sensor fusion tasks. To process continuous-valued input, \glspl{snn} require encoding schemes that convert signals into spike trains \cite{guo_neural_2021}. 

Efficient deployment of \glspl{snn} on embedded hardware platforms poses additional complexity. Conventional hardware accelerators, optimized for dense numerical operations and synchronous computation, are ill-suited to the sparse, event-driven, and temporally dynamic nature of \glspl{snn} \cite{10.1145/3571155}. As a result, specialized neuromorphic hardware accelerators have been developed to exploit these unique characteristics. Examples include IBM’s TrueNorth ASIC \cite{akopyan_truenorth_2015}, which implements large-scale digital \glspl{snn} with extremely low power consumption, and Intel’s Loihi processor \cite{davies_loihi_2018}, which supports on-chip learning and programmable synaptic plasticity. While these platforms demonstrate the feasibility of energy-efficient spike-based computation, they often lack integration flexibility and impose constraints on network structure and learning rules. Alternatively, \gls{fpga}-based approaches have emerged as adaptable solutions capable of emulating diverse \gls{snn} models with customized architectures~\cite{neil_minitaur_2014,liu_low_2023,nevarez_accelerating_2021,li_firefly_2023,gerlinghoff_e3ne_2022,khodamoradi_s2n2_2021, li_fast_2021}. However, existing implementations frequently involve significant manual design effort, lack modularity, or are tightly coupled to specific toolchains and hardware configurations, hindering their usability and scalability.

In this paper, we present our contribution to the Digit Recognition Low Power and Speed Challenge @ ICIP 2025\footnote{\url{https://mlunglma.github.io/challenge.html}}, which focuses on evaluating \gls{fpga}-based accelerators in terms of inference speed, energy efficiency, and classification accuracy on the MNIST dataset \cite{deng_mnist_2012}. Our approach is centered around \texttt{Spiker+} \cite{10794606}, an open-source, end-to-end framework for the design, training, and deployment of \glspl{snn} on \gls{fpga} platforms. Unlike traditional fixed-function accelerators, \texttt{Spiker+} enables the automatic generation of optimized hardware tailored to specific tasks. This flexibility allows us to synthesize an accelerator specifically optimized for handwritten digit recognition, considering the trade-offs between speed, power, and accuracy.

The framework supports high-level \gls{snn} modeling in Python, integrates surrogate-gradient-based training through \texttt{snnTorch} \cite{eshraghian_training_2023}, and performs automatic quantization to match hardware constraints. It then generates synthesizable HDL, streamlining deployment on target \gls{fpga} platforms. This seamless integration of training and hardware generation enables rapid design-space exploration and functional prototyping, also supported by companion tools such as SpikeExplorer \cite{electronics13091744} and \texttt{Optuna} \cite{akiba2019optuna}. In the context of the challenge, we demonstrate that using \texttt{Spiker+} we were able to effectively co-optimize both the neural network architecture and its hardware implementation, achieving high accuracy while meeting strict energy and latency requirements. Our results validate the viability of task-specific, reconfigurable \gls{snn} accelerators for low-power edge image processing applications.

\section{Methods}
\label{sec:methods}

\begin{figure*}[t]
    \centering
    \includegraphics[width=\textwidth]{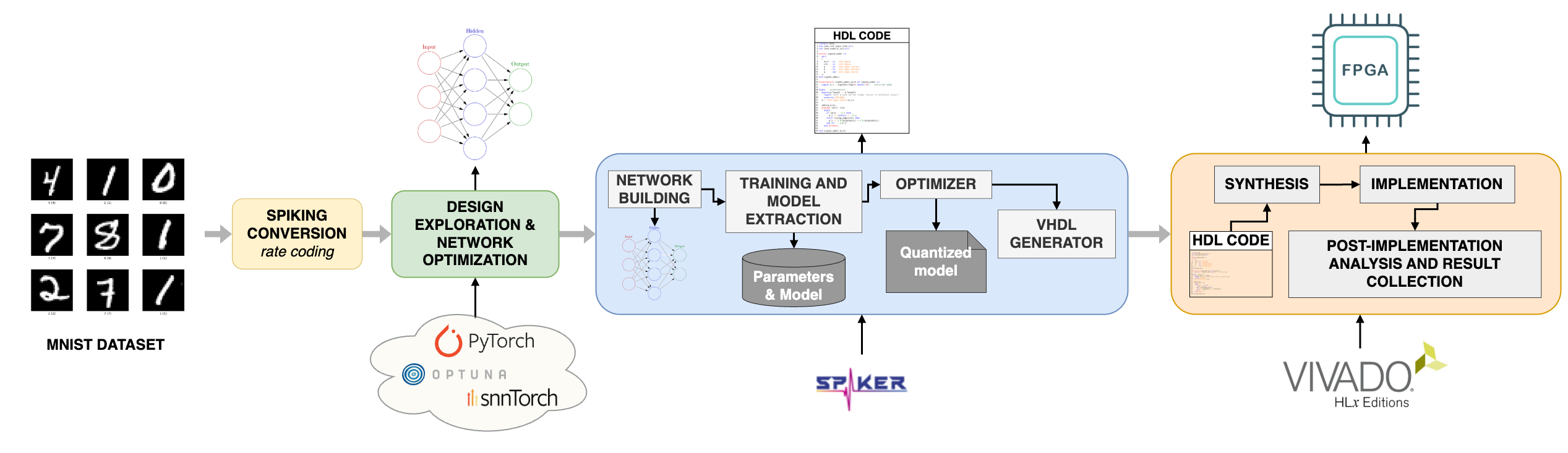}
    \caption{Workflow scheme}
    \label{fig:flow}
\end{figure*}

\autoref{fig:flow} sketches the entire flow required to start from the MNIST database and to obtain a highly optimized \gls{snn} accelerator for digit recognition to be deployed on \glspl{fpga} through the use of the \texttt{Spiker+} framework, an open-source framework, publicly available on GitHub. 
To facilitate full reproducibility of the experiments presented in this paper, we prepared a dedicated repository containing detailed implementation instructions and supplementary materials. This repository is accessible at \url{https://github.com/smilies-polito/sfatti}

\subsection{Spike Encoding}
When processing non-neuromorphic datasets such as MNIST, \glspl{snn} require an input encoding stage to transform static, continuous-valued data into spike trains. This transformation is necessary to interface conventional data formats with the discrete, event-driven dynamics of spiking neurons.
Rate coding is the most widely adopted encoding scheme when dealing with static data, such as images \cite{guo_neural_2021}. In rate coding, pixel magnitude is mapped to spike frequency within a fixed temporal window, offering robustness and simplicity. In our architecture, input encoding is handled via Poisson-based rate coding, realized using the \texttt{snntorch.spikegen} module. Each pixel intensity is interpreted as a firing probability at every timestep over a window of a fixed number of steps. This encoding procedure is applied before submitting the images to the accelerator and is also integrated into the training pipeline through a dedicated preprocessing class.
Importantly, no additional encoding is required for inherently event-based inputs, such as those from neuromorphic sensors like \gls{dvs} cameras, as the data are already represented in spike form.

\subsection{Network Training}

Training \glspl{snn} presents notable challenges primarily due to the discrete, non-differentiable nature of spike generation. This characteristic violates the assumptions underlying standard gradient descent methods, such as backpropagation, which rely on the availability of continuous and differentiable activation functions \cite{eshraghian2023training}. Consequently, direct application of gradient-based optimization to \glspl{snn} is infeasible. To address this, surrogate gradient methods have been introduced \cite{neftci2019surrogate}. These methods approximate the discontinuous spike activation functions with smooth, differentiable surrogates during the backward pass of \gls{bptt} \cite{werbos1990backpropagation}, while preserving the original spike behavior during the forward pass \cite{wu2018spatio}. This enables effective supervised training of \glspl{snn} using conventional optimization algorithms and loss functions. Recent advances have demonstrated that surrogate-gradient training can achieve performance comparable to traditional \glspl{ann} on standard benchmarks \cite{li2024brain}, closing the gap between bio-plausibility and practical efficacy.
In our work, the training phase is handled entirely in software using the \texttt{SnnTorch} framework \cite{eshraghian2023training}, which is fully integrated with \texttt{Spiker+}. \texttt{SnnTorch} supports surrogate-gradient learning, specifically \gls{bptt} with differentiable approximations of the spike function, and is used to optimize all synaptic weights and neuron thresholds in floating-point precision on a host CPU or GPU. 

To ensure highly efficient hardware acceleration, \texttt{Spiker+} imposes a hardware-friendly constraint on the neuron model used during training. Specifically, \texttt{Spiker+} employs a standard \gls{lif} neuron model, where the membrane potential decays exponentially over time. This decay is governed by two parameters, $\alpha$ and $\beta$, which control the rate at which the potential and synaptic current diminish. In traditional implementations, these exponential decays would require multiplication operations, which are costly regarding hardware resources. To eliminate the need for multipliers and simplify the digital logic, \texttt{Spiker+} requires rounding these parameters to the nearest power of two during training. This allows the corresponding decay operations to be implemented as simple binary shifts on the \gls{fpga}, significantly reducing area and power consumption.
While this approximation introduces a slight degradation in classification accuracy, the trade-off is justified by the substantial improvements in hardware efficiency. Once training is completed with these quantized decay parameters, the learned model is passed to the quantization module for further refinement and synthesis, ensuring that the final implementation meets the design constraints of the target \gls{fpga} platform.

To ensure reproducibility across environments, we fixed the random seed for all relevant components, including \texttt{Python}, \texttt{NumPy}, and \texttt{PyTorch}. This guarantees consistent behavior across runs, essential for two reasons: the spike encoding process uses Poisson-based rate coding, introducing randomness in input generation; and training involves stochastic operations such as data shuffling and weight initialization. Even though the seed is fixed, results may still vary slightly due to nondeterministic operations in underlying libraries or differences in hardware backends. As highlighted in the PyTorch documentation \cite{pytorch_reproducibility}, perfect reproducibility across all platforms and configurations cannot be guaranteed.

\begin{table*}[ht]
    \centering
    \scriptsize
    \renewcommand{\arraystretch}{1.2}
    \setlength{\tabcolsep}{4pt}
    \caption{Summary of \glspl{snn} configurations and hardware evaluation results. The network architecture summarizes the number of neurons per layer. The architecture is a feed-forward, fully connected network. Quantization is shown as weights bit-width (WB), membrane bit-width (MB), and fixed-point decimal bits (FPd). Utilization is shown as a percentage of used resources with respect to the total of \glspl{lut}, \glspl{ff}, and \glspl{bram}}
    \begin{tabular}{|c|c|ccc|c|c|c|ccc|c|c|}
        \hline
        \textbf{Neuron} & \textbf{Model} &
        \multicolumn{3}{c|}{\textbf{Quantization}} &
        \textbf{Accuracy (\%)} & \textbf{Power (mW)} & \textbf{Timesteps} &
        \multicolumn{3}{c|}{\textbf{Utilization (\%)}} &
        \textbf{Clock (MHz)} & \textbf{(Img/s)/W} \\
        \cline{3-5} \cline{9-11}
        & & \textbf{WB} & \textbf{MB} & \textbf{FPd} & & &  & \textbf{LUTs} & \textbf{FFs} & \textbf{BRAMs} & & \\
        \hline
        \multirow{2}{*}{I-order LIF} & \multirow{2}{*}{784-25-10} & 4 & 4 & 4 & 73.50 & 162 & 10 & 1.92 & 0.95 & 1.23 & 188.6 & / \\
         &  & 10 & 10 & 6 & 96.19 & 187 & 10 & 2.44 & 1.05 & 2.62 & 166.6 & / \\
        \hline
        \multirow{2}{*}{I-order LIF} & \multirow{2}{*}{784-50-10} & 4 & 4 & 4 & 83.29 & 177 & 10 & 2.31 & 1.06 & 2.15 & 188.6 & / \\
         &  & 10 & 10 & 6 & 97.16 & 222 & 10 & 3.28 & 1.24 & 4.77 & 161.2 & / \\
        \hline
        \multirow{3}{*}{I-order LIF} & \multirow{3}{*}{784-75-10} & 4 & 4 & 4 & 86.26 & 191 & 10 & 2.74 & 1.17 & 2.92 & 185.2 & / \\
        \rowcolor{gray!15}
        I-order LIF & 784-75-10 & 6 & 9 & 5 & 97.54 & 231 & 10 & 3.93 & 1.38 & 4.15 & 163.9 & 81,712.5 \\
         &  & 10 & 10 & 6 & 97.86 & 254 & 10 & 4.13 & 1.43 & 6.92 & 153.8 & 69,692.9 \\
        \hline
        \multirow{3}{*}{I-order LIF} & \multirow{3}{*}{784-100-10} & 4 & 4 & 4 & 73.53 & 202 & 10 & 3.15 & 1.28 & 3.85 & 172.4 & / \\
         &  & 6 & 9 & 5 & 97.59 & 251 & 10 & 4.71 & 1.56 & 5.54 & 156.2 & 71,696.4 \\
         &  & 10 & 10 & 4 & 97.78 & 285 & 10 & 4.95 & 1.61 & 9.08 & 151.5 & 61,227.4 \\
        \hline
    \end{tabular}
    \label{tab:snn_results_improved_multicol}
\end{table*}

\subsection{High-Level Design Space Exploration}

Before committing to hardware synthesis, we performed a high-level exploration of the design space using purely software-based tools. This phase relied on \texttt{SnnTorch} for training and inference, and \texttt{Optuna} \cite{akiba2019optuna} for large-scale hyperparameter optimization.
This initial analysis focused on identifying promising network configurations from an algorithmic standpoint. Various architectures, differing in neuron types, layer sizes, spiking conversion time-steps, and other training-specific parameters, were evaluated to assess their accuracy, trying to minimize those parameters that influence power and latency. The goal was to rapidly discard configurations that failed to meet target requirements through a computationally efficient approach that aimed at surveying the design space and identifying viable candidate networks.


\subsection{Quantization Exploration with Spiker+}

Following the high-level screening, we refined the design space using the quantization and hardware-aware simulation capabilities of \texttt{Spiker+}. The framework includes a quantization optimizer that reduces memory and computational costs by exploring efficient numerical formats for weights, membrane potentials, and fractional precision. It uses signed two’s complement fixed-point arithmetic with overflow protection via value clipping. The optimizer accepts search ranges for bit widths (e.g., weights: 4–10 bits) and performs a combinatorial sweep, evaluating each configuration by running post-quantization inference and recording the resulting accuracy.

Along with the standard exploration used by \texttt{Spiker+}, we modified the framework to enable automated batch evaluations, logging all configurations and corresponding results into structured files. This eliminated the need for manual selection and ensured reproducibility, together with a faster study of the behaviour of many configurations. For each viable network architecture (e.g., fixed neuron model, variable layer sizes), we executed a full set of quantized simulations and applied application-specific constraints, such as a minimum test accuracy of 97.5\%, to filter the results. This approach enabled a comprehensive yet automated exploration of trade-offs between accuracy, latency, and hardware cost, significantly accelerating the design of efficient and performant \gls{snn} accelerators.

\subsection{HDL Generation}

After optimizing all design parameters, \texttt{Spiker+} is used to generate a synthesizable hardware description of the target accelerator. To achieve this goal, \texttt{Spiker+} accepts the network architecture and trained and quantized network parameters, producing \gls{vhdl} source code describing the complete accelerator. This includes neuron modules, synaptic memory blocks, control logic, and I/O interfaces. \texttt{Spiker+} supports a full II-order \gls{lif} neuron model, where both $\alpha$ and $\beta$ are non-zero. However, by adjusting the neuron parameters, it is also possible to implement simplified I-order \gls{lif} neurons ($\alpha=0$), or even basic \gls{if} neurons by setting both $\alpha$ and $\beta$ to zero. 

\texttt{Spiker+} supports two generation modes: (i) simulation and verification, and (ii) direct deployment to physical \glspl{fpga}. In our case, we selected the deployable mode, which configures the memory blocks to match the on-chip \gls{bram} of the target \gls{fpga} and sets all timing parameters for real-time operation. As mentioned before, the generated architecture avoids multipliers by approximating exponential decays with bit-shift operations (enabled by the power-of-two rounding of $\alpha$, $\beta$), and processes input spikes sequentially to mitigate bandwidth and memory limitations typical of low-end \glspl{fpga}. This results in compact and power-efficient implementations, as demonstrated in our evaluation.

\subsection{Implementation and Results Gathering}

The VHDL project generated by \texttt{Spiker+} was integrated into a Vivado project . The desgined was then synthesized and mapped to Xilinx Kintex XC7K160TFBG484-1 \gls{fpga} \cite{amd_kintex7}. The final deployment requires two key steps. While \texttt{Spiker+} automatically instantiates all required \glspl{rom} in the design, the user is responsible for initializing their content. This is achieved through \texttt{.coe} files generated by \texttt{Spiker+}, which must be loaded into the memory generator \glspl{ip} within the Vivado design environment. Additionally, to correctly map the accelerator’s I/O interfaces to the physical pins of the \gls{fpga}, the user must supply a custom \texttt{XDC} file. This file defines the clock and reset sources, along with the placement of data and control signals on the target device.
After synthesizing and implementing the design, the bitstream is loaded onto the \gls{fpga}. During runtime, key performance metrics such as maximum clock speed, power consumption, and inference speed are collected. These metrics are correlated with the \gls{dse} results to validate the effectiveness of the automated design methodology.

\section{Results}
\label{sec:res}

\autoref{tab:snn_results_improved_multicol} summarizes the \gls{dse} conducted to implement efficient \gls{snn}-based digit recognition on \gls{fpga}. The row highlighted in gray indicates the architecture with the highest energy efficiency, measured in images per second per watt, which was ultimately selected as our candidate design for the challenge.

All models were trained using surrogate gradient descent; each underwent training for 25 epochs using a batch size of 64 and the Adam optimizer with a learning rate of $5 \cdot 10^{-4}$ and hyperparameters $\beta_1 = 0.9$ and $\beta_2 = 0.999$. The training objective was defined by the cross-entropy loss function. Training was performed in floating-point precision using the MNIST training set (60,000 samples), followed by post-training quantization inference to assess post-training accuracy. The final accuracy was measured via a bit-accurate software simulation of the accelerator using the standard MNIST test set (10,000 samples). To ensure consistency and reproducibility, the dataset was sourced from the official \texttt{torchvision.datasets} module.

Latency was assessed through hardware-level simulation of the quantized \gls{snn} architecture. Specifically, we measured the number of clock cycles required to perform inference on a single MNIST sample; this metric depends on the temporal encoding window (i.e., the number of timesteps). 
To estimate the number of images processed per second (img/s), we analyzed the results from the functional simulation. A parallel input interface was employed in the accelerator, with one input line per neuron. Since inputs are processed sequentially, reduced parallelism can be supported by adopting a dual-buffer interface: the two buffers alternate in supplying inputs to the accelerator. While one buffer is actively feeding the computation, the other is loaded with new data, ensuring continuous operation without introducing any latency overhead.

The latency estimation starts measuring during simulation the time $\Delta t$ required by the accelerator to process a single input sample as defined in (\ref{eq:delta_t}). The beginning of the inference is marked rising the \texttt{start} signal in the accelerator and the end is detected with an active value on the \texttt{ready} signal. This measure is done with a standard clock cycle of 20ns defined by Spiker+ in its testbench instead of the target clock cycle identified post synthesis for the best target architecture (see Table \ref{tab:snn_results_improved_multicol}).
\vspace{-12pt}
\begin{equation}
    {\Delta t = t_{\text{end}} - t_{\text{start}} = (347{,}510 - 173{,}810)\,\text{ns} = 173{,}700\,\text{ns}}
    \label{eq:delta_t}
\end{equation}

The throughput $R$ can be computed as the inverse of this processing time according to (\ref{eq:throughput}):
\begin{equation}
    R = \frac{1}{\Delta t} = \frac{1}{173{,}700 \times 10^{-9}} \approx 5{,}757\ \text{img/s}
    \label{eq:throughput}
\end{equation}

Eventually, the number of clock cycles required to complete a single inference can be precisely determined by dividing the total inference time by the clock period, as shown in (\ref{eq:cycles_count}). 

\begin{equation}
    {\#Cycles = \displaystyle \frac{173{,}700\,\text{ns}}{20\,\text{ns}} = 8{,}685\ \text{cycles}}
    \label{eq:cycles_count}
\end{equation}

Since this cycle count remains constant across all network configurations, the latency for each configuration can be derived by simply multiplying the number of cycles by the architecture's clock period. This methodology yields a realistic throughput estimation under timing-accurate conditions, as it captures the actual temporal behavior observed during simulation.
The result illustrates, for a representative sample, the achieved image processing throughput of the accelerator, which is approximately 18{,}875 images per second under the given timing conditions (\ref{eq:final_count}).

\begin{align}
    \Delta t &= 8{,}685\ \text{cycles} \cdot 6{.}1\ ns = 52{,}978{.}5\ ns \nonumber \\
    R &= \frac{1}{\Delta t} = \frac{1}{52{,}978{.}5 \times 10^{-9}} = 18{,}875{.}6\ \text{img/s}
    \label{eq:final_count}
\end{align}

To estimate hardware costs and performance, each configuration was synthesized for the selected \gls{fpga} \cite{amd_kintex7} using the Vivado tool, and post-synthesis reports were collected. Dynamic power consumption was estimated assuming a default toggle rate of 12.5\% at the maximum achievable clock frequency, determined via timing analysis in Vivado. Architectures using reduced numerical precision and smaller hidden layers demonstrated lower power consumption with minimal impact on accuracy.
Resource utilization was analyzed in terms of \glspl{lut}, flip-flops, and \glspl{bram} (no DSP blocks are used). All tested designs fit comfortably within the resource constraints of the target device, leaving sufficient headroom for integration into complete systems.

\section{Conclusions}
\label{sec:conclusion}

This work presents an end-to-end Spiker+-based flow that converts high-level SNNs into deployable FPGA accelerators while systematically stripping away non-essential blocks to meet the low-power mandate of the Grand Challenge. The flow abstracts much of the manual effort traditionally required to bring SNNs onto resource-constrained devices. Beyond the convenience factor, the approach highlights how temporal information encoding fundamentally reshapes accelerator design: instead of a single, synchronous pass over dense tensors, computation unfolds over multiple timesteps, with sparse spike events. However, custom datapaths, memory tiling and efficient data conversion compress the effective decision window, keeping end-to-end latency competitive with ANN accelerators while lowering dynamic energy per inference.

Looking beyond the competition, the same flow can re-enable the deferred features—on-chip learning, adaptive neurons, dynamic partial reconfiguration—and extend to fully event-based datasets. These additions are orthogonal to the challenge rules yet pivotal for future edge platforms that must adapt online, share resources and operate under stricter energy envelopes. Taken together, the methodology and its planned extensions narrow the gap between neuromorphic theory and deployable, power-efficient hardware, underscoring SNNs as a viable alternative to conventional networks when both energy and responsiveness matter

\bibliographystyle{IEEEbib}


\end{document}